\documentclass{INTERSPEECH2023}

\interspeechcameraready 

\usepackage{times}
\usepackage{latexsym}
\usepackage{caption}
\usepackage{graphicx}
\usepackage{latexsym}
\usepackage{booktabs}
\usepackage{xcolor,colortbl}
\usepackage{amsmath}
\usepackage{multirow}
\usepackage[normalem]{ulem}
\usepackage{lipsum}
\usepackage{colortbl}
\usepackage{soul}
\usepackage{hyperref}

\setul{0.5ex}{0.2ex}
\newcommand{\ulcolor}[2][]{\setulcolor{#1}\ul{#2}}
\definecolor{darkgreen}{RGB}{4,182,117}

\title{Cross-Lingual Cross-Age Group Adaptation\\for Low-Resource Elderly Speech Emotion Recognition}
\name{Samuel Cahyawijaya$^*$\thanks{$^*$ Equal contribution.}, Holy Lovenia$^*$, Willy Chung$^*$, Rita Frieske, Zihan Liu, Pascale Fung}
\address{
  The Hong Kong University of Science and Technology}
\email{\{scahyawijaya, hlovenia, whcchung\}@connect.ust.hk}

\begin{document}

\maketitle
 
\begin{abstract}

Speech emotion recognition plays a crucial role in human-computer interactions. However, most speech emotion recognition research is biased toward English-speaking adults, which hinders its applicability to other demographic groups in different languages and age groups. In this work, we analyze the transferability of emotion recognition across three different languages--English, Mandarin Chinese, and Cantonese; and 2 different age groups--adults and the elderly. To conduct the experiment, we develop an English-Mandarin speech emotion benchmark for adults and the elderly, BiMotion, and a Cantonese speech emotion dataset, YueMotion. This study concludes that different language and age groups require specific speech features, thus making cross-lingual inference an unsuitable method. However, cross-group data augmentation is still beneficial to regularize the model, with linguistic distance being a significant influence on cross-lingual transferability. We release publicly release our code at \url{https://github.com/HLTCHKUST/elderly_ser}.

\end{abstract}

\section{Introduction}

Understanding human emotion
is a crucial step towards better human-computer interaction~\cite{beale2008hci,cowie2001emot-rec,winata2021nora,ishii2021erica}. 
Most studies on speech emotion recognition are centered on young-adult people, mainly originating from English-speaking countries~\cite{busso2008iemocap,livingstone2018ryerson,zadeh2018multi,poria2019meld}. This demographic bias causes existing commercial emotion recognition systems to inaccurately perceived emotion in the elderly~\cite{kim2021agebias}.
Despite the fast-growing elderly demographic in many countries~\cite{rouzet2019ageing},
only a few studies with a fairly limited amount of data work on emotion recognition for elderly~\cite{cao2014crema,ma2019elderreact,pichora2020tess}, especially from non-English-speaking countries~\cite{lee2017csed}.

To cope with the limited resource problem in the elderly emotion recognition, in this work, we study the prospect of transferring emotion recognition ability over various age groups and languages through the utilization of multilingual pre-trained speech models, e.g., Wav2Vec 2.0~\cite{baevski2020wav2vec2}. Specifically, we aim to understand the transferability of emotion recognition ability using only speech modality between two languages and two age groups, i.e, English-speaking adults, English-speaking elderly, Mandarin-speaking adults, and Mandarin-speaking elderly. To do this, we develop BiMotion, a bi-lingual bi-age-group speech emotion recognition benchmark that covers 6 adult and elderly speech emotion recognition datasets from English and Mandarin Chinese. Additionally, we analyze the effect of language distance on the transferability of emotion recognition ability from Mandarin and English using YueMotion, a newly-constructed Cantonese speech emotion recognition dataset.

\begin{figure}[t!]
    \centering
    \resizebox{0.8\linewidth}{!}{
        \includegraphics{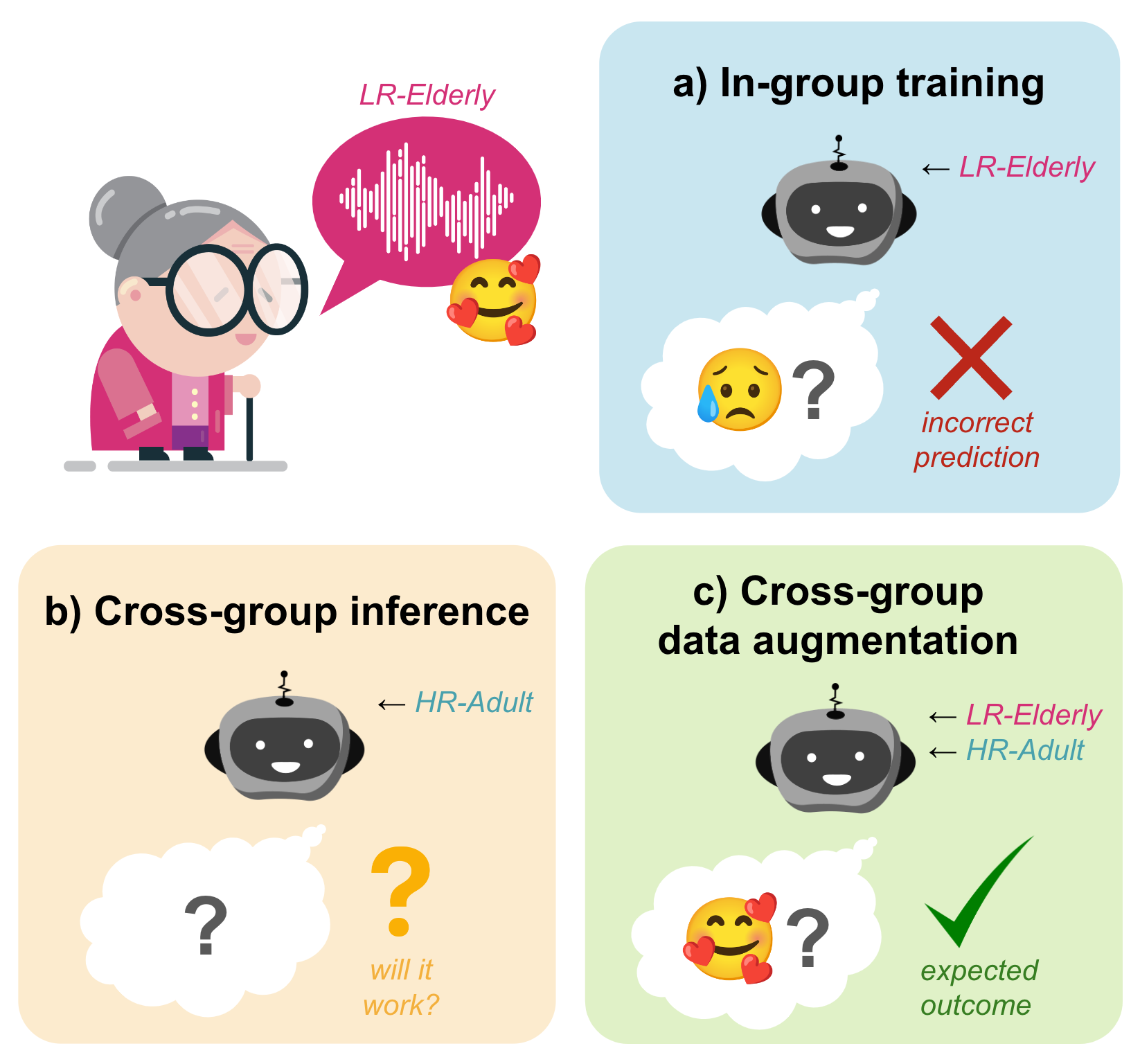}
    }
    \caption{Elderly speech emotion data in low-resource languages is extremely rare, which makes an in-group trained model perform poorly (a). We explore two ways to improve it: cross-group inference (b) and cross-group augmentation (c).}
    \label{fig:my_label}
    \vspace{-15.5pt}
\end{figure}

We analyze the transferability of emotion recognition ability in three ways, i.e., 1) cross-group inference using only the source group data for training to better understand the features distinction between each group, 2) cross-group data augmentation to better understand the transferability across different groups, and 3) feature-space projection to better understand the effect of transferability across different language distance. Through a series of experiments, we conclude that:
\begin{itemize}
    \item Very distinct speech features are needed for recognizing emotion across different language and age groups, which makes cross-lingual inference an ill-suited method for transferring speech emotion recognition ability.
    \item Despite the different important speech features across groups, data augmentation across different groups is still beneficial as it helps the model to better regularize yielding higher evaluation performance.
    \item Language distance plays a crucial role in cross-lingual transferability, as more similar languages tend to share more similar speech characteristics, thus allowing better generalization to the target language.
\end{itemize}



\section{Related Work}

\subsection{Emotion Recognition}
Various efforts have explored emotion recognition solutions for around two decades ago through different modalities~\cite{martnez2005emot-ai,bertero2016multimodalhumor,cen2010speechemot,bertero2017speechemot,lovenia2022did,latif2022aiemot,dai2021weaklysupervised,dai2021mm-emot,lubis2014construction,lubis2016emotion}. Most studies focus on methods that can estimate the subject emotion effectively through a specific modality while some others focus on combining multiple modalities to better estimate the subject emotion. 
The transferability of emotion recognition across different cultures and language groups has also been previously studied~\cite{hareli2015crosscultural,lim2016cultural,iosifov2022serlangtrans} showing that there are significant differences across languages and cultures.
Nevertheless, it is only evaluated on a small-scale model that owns limited world knowledge. In this work, we extend this analysis to pre-trained speech models and further explore the transferability to cover an extremely low-resource group, i.e., the elderly in the low-resource language group.


\subsection{Pre-trained Speech Model} 
Large pre-trained speech models achieve state-of-the-art performance for various speech tasks in recent years~\cite{baevski2020wav2vec2,hsu2021hubert,babu2022interspeech}.
These models have been trained in an unsupervised manner on large-scale multilingual speech corpora. The representation of these models has been shown to be effective, achieving state-of-the-art performance in various downstream tasks including automatic speech recognition, speaker verification, language identification, and emotion recognition~\cite{fan2021verification-lid,pepino2021emotionrecognition,lovenia2022ascend,cahyawijaya2022nusacrowd,dai2022ci}. In this work, we adopt the XLSR Wav2Vec 2.0 model~\cite{baevski2020wav2vec2,babu2022interspeech}, which has been pre-trained on large-scale multilingual speech corpora. The multilingual representation learned by the model will be crucial for evaluating the transferability of emotion recognition ability across different languages.

\section{Benchmark and Dataset}

\subsection{BiMotion Benchmark}

\begin{table}[!t]
    \caption{Statistics of datasets covered in BiMotion.}
    \label{tab:bimotion-data}
    \centering
    \resizebox{0.87\linewidth}{!}{
    \begin{tabular}{l|c|c|c|c|c}
        \toprule
        \textbf{Dataset} & \textbf{Language} & \textbf{Age Group} & \textbf{\#Train} & \textbf{\#Valid} & \textbf{\#Test} \\ \midrule
        \multirow{2}{*}{\textbf{CREMA-D}} & English & Elderly & 150 & 42 & 300 \\
        & English & Adults & 5000 & 750 & 1200 \\ \midrule
        \textbf{CSED} & Mandarin & Elderly & 200 & 52 & 400 \\ \midrule
        \textbf{ElderReact} & English & Elderly & 615 & 355 & 353 \\ \midrule
        \multirow{2}{*}{\textbf{ESD}} & Mandarin & Adults & 15000 & 1000 & 1500 \\
        & English & Adults & 15000 & 1000 & 1500 \\ \midrule
        \textbf{IEMOCAP} & English & Adults & 7500 & 1039 & 1500 \\ \midrule
        \multirow{2}{*}{\textbf{TESS}} & English & Elderly & 699 & 200 & 500 \\
        & English & Adults & 700 & 201 & 500 \\
        \bottomrule
    \end{tabular}
    }
\end{table}

To test the transferability of emotion recognition ability across different languages and age groups, we develop a bilingual speech emotion recognition benchmark namely BiMotion, that covers two languages, i.e., English and Mandarin, and two age groups, i.e., Elderly (age $\geq$60 y.o) and Adults (age 20-59). We construct our benchmark from a collection of six publicly available datasets, i.e., CREMA-D~\cite{cao2014crema}, CSED~\cite{lee2017csed}, ElderReact~\cite{ma2019elderreact}, ESD~\cite{zhou2022esd}, IEMOCAP~\cite{busso2008iemocap}, and TESS~\cite{pichora2020tess}. For datasets with an official split, i.e., ESD and ElderReact, we use the provided dataset splits. For the other datasets, we generate a new dataset split since there is no official split provided. The statistics of the datasets in BiMotion are shown in Table~\ref{tab:bimotion-data}.

\subsection{YueMotion Dataset}

To strengthen our transferability analysis, we further introduce a new speech emotion recognition dataset covering adults in Cantonese named YueMotion. The utterances in YueMotion are recorded by 11 speakers (4 males and 7 females) each with a personal recording device. During the recording session, each speaker is asked to say out loud 10 pre-defined sentences used in daily conversations with 6 different emotions, i.e., happiness, sadness, neutral, disgust, fear, and anger. The YueMotion dataset is used to further verify our hypothesis regarding the transferability of emotion recognition ability. The detailed statistics of YueMotion dataset are shown in Table~\ref{tab:yuemotion-data}.

\begin{table}[]
    \caption{Statistics of the YueMotion dataset.}
    \label{tab:yuemotion-data}
    \centering
    \resizebox{0.85\linewidth}{!}{
    \begin{tabular}{c|c|c|c|c|c}
        \toprule
        \textbf{Age Group} & \textbf{Gender} & \textbf{\#Train} & \textbf{\#Valid} & \textbf{\#Test} & \textbf{\#All} \\ \midrule
        \multirow{2}{*}{Adults} & Male & 120 & 36 & 84 & 240 \\
        & Female & 210 & 63 & 147 & 420 \\ \midrule
        \multicolumn{2}{c|}{\textbf{Total Samples}} & \textbf{330} & \textbf{99} & \textbf{231} & \textbf{660} \\
        \bottomrule
    \end{tabular}
    }
\end{table}

\section{Methodology}

We employ two methods to evaluate the transferability of emotion recognition ability, i.e., \ulcolor[darkgreen]{cross-group data augmentation} and \ulcolor[orange]{cross-group inference}.\footnote{To enhance readability, we use colored underlines to mark the term \ulcolor[darkgreen]{cross-group data augmentation} and \ulcolor[orange]{cross-group inference} onwards.} The detail of each method is explained in the following subsection.

\subsection{Cross-Group Inference}
\label{sec:cross-group-inf}

We define a set of language L=\{$l_1$, $l_2$, \dots, $l_n$\} and a set of age group A=\{$a_1$, $a_2$, \dots, $a_m$\}. We define training and evaluation data from a language $l_i$ and age group $a_j$ as $X^{trn}_{l_i, a_j}$ and $X^{tst}_{l_i, a_j}$, respectively.
To understand the transferability from the source group to the target group, we explore a method called cross-group inference. Following prior works on zero-shot accent/language adaptation~\cite{winata2020crossaccent,hu2020xtreme,winata2022nusax}, we explore cross-lingual and cross-age inference settings in our experiment. Specifically, given a training dataset $X^{trn}_{l_i,a_j}$, we select three out-of-group test datasets: 1) datasets that are not in $l_i$ language ($X^{tst}_{\neg l_i,a_j}$), 2) datasets that are not in $a_j$ age-group ($X^{tst}_{l_i,\neg a_j}$), and 3) datasets that are not in $l_i$ language and $a_j$ age-group $X^{tst}_{\neg l_i, \neg a_j}$. We conduct cross-group inference by training the models on a specific group training set and evaluating them on the three out-of-group test sets. With this method, we can analyze the effect of cross-group information on a specific language and age group.

\subsection{Cross-Group Data Augmentation}
\label{sec:cross-group-da}

Prior works~\cite{huang2018auggan,singh2019xlda} have shown that cross-domain and cross-lingual data augmentation method effectively improves the model performance in textual and speech modalities, especially on low-resource languages and domains. We adopt the data augmentation technique to the speech emotion recognition task by utilizing cross-lingual and cross-age data augmentation. Our cross-group data augmentation injects a training dataset $X^{trn}_{l_i, a_j}$ from a language group $l_i$ and an age group $a_j$ with data from other language and age groups producing a new augmented dataset. Given $X^{trn}_{l_i, a_j}$, there are three different data augmentation settings possible, i.e., cross-lingual data augmentation on the same age group resulting in $X^{trn}_{L,a_j}$, cross-age data augmentation on the same language resulting in $X^{trn}_{l_i,A}$, and cross-lingual cross-age data augmentation resulting in $X^{trn}_{L,A}$. We then fine-tune the model on the augmented dataset and evaluate it on $X^{tst}_{l_i, a_j}$, the test set with a language $l_i$ and age group $a_j$. Through this method, we can analyze the influence of cross-group augmentation on a specific language and age group.

\subsection{Feature-Space Projection}
\label{sec:feature-projection}

Given a fine-tuned emotion recognition model $\theta$, we extract the speech representation by taking the high-dimensional features before the last linear classification layer. By using $\theta$, we extract the speech features from data points and perform matrix decomposition~\cite{jolliffe1986pca,hyvarinen2000ica,halko2011md,lovenia2019td,winata2020lr,cahyawijaya2021greenformer,cahyawijaya2021greenformers} into a 2-dimensional space with UMAP~\cite{mcinnes2018umap-software} for visualization. With this approach, we can approximate the effect of \ulcolor[darkgreen]{augmenting cross-group data points} to the training coverage, which will determine the prediction quality of the model to the evaluation data in a specific language $l_i$ and age-group $a_j$.

\section{Experimental Settings}

\begin{table}[!t]
    \caption{Evaluation results of \ulcolor[orange]{cross-lingual and cross-age groups inference} on BiMotion. \textbf{Cross-all} denotes cross-age cross-lingual inference. \textbf{eng}, \textbf{zho}, \textbf{eld}, and \textbf{adu} denote English, Mandarin, elderly, and adults, respectively.}
    \label{tab:cgi-result}
    \centering
    \resizebox{\linewidth}{!}{
        \begin{tabular}{c|c|c|c|c|c}
        \toprule
        \multirow{3}{*}[-0.3em]{\bf{\begin{tabular}[c]{@{}c@{}}Training\\Data\end{tabular}}} & \multicolumn{4}{c|}{\bf{Test Data}} & \multirow{3}{*}{\bf{Avg.}} \\ \cmidrule{2-5}
         & $\bm{l_i}$\bf{=eng} & $\bm{l_i}$\bf{=eng} & $\bm{l_i}$\bf{=zho} & $\bm{l_i}$\bf{=zho} & \\
         & $\bm{a_j}$\textbf{=eld} & $\bm{a_j}$\textbf{=adu} & $\bm{a_j}$\textbf{=eld} & $\bm{a_j}$\textbf{=adu} & \\ \midrule
        \textbf{Cross-all ($\bm{X^{trn}_{\neg l_i, \neg a_j}}$)} & 27.48 & 9.71 & 24.75 & 28.71 & 22.66 \\
        \textbf{Cross-age ($\bm{X^{trn}_{l_i, \neg a_j}}$)} & 40.40 & 27.69 & 26.69 & 14.56 & 27.34 \\
        \textbf{Cross-lingual ($\bm{X^{trn}_{\neg l_i, a_j}}$)} & 14.54 & 32.42 & 16.06 & 57.54 & 30.14 \\
        \textbf{Baseline ($\bm{X^{trn}_{l_i, a_j}}$)} & \bf{55.12} & \bf{70.28} & \bf{57.00} & \bf{97.00} & \bf{69.85} \\
        \bottomrule
        \end{tabular}
    }
\end{table}

\subsection{Baseline Models}

For our experiment, we utilize the XLSR-53 Wav2Vec 2.0~\cite{baevski2020wav2vec2,babu2022interspeech} model which is pre-trained in 53 languages including English, Mandarin, and Cantonese.\footnote{We use the model checkpoint from \url{https://huggingface.co/facebook/wav2vec2-large-xlsr-53}.} We compare the performance of the \ulcolor[darkgreen]{cross-group data augmentation} and \ulcolor[orange]{cross-group inference} settings with a simple in-group training baseline where given a language $l_i$ and an age-group $a_j$, the model is fine-tuned only on $X_{l_i, a_j}$ and evaluated on the test set of the corresponding language and age-group.

\subsection{Training Settings}

To evaluate the transferability of emotion recognition ability, we conduct an extensive analysis in the BiMotion benchmark. We explore two different cross-group transfer methods, i.e., cross-group data-augmentation (\S\ref{sec:cross-group-da}) and cross-group inference (\S\ref{sec:cross-group-inf}). Specifically, we fine-tune the pre-trained XLSR-53 Wav2Vec 2.0 model using the in-group and the \ulcolor[darkgreen]{cross-group augmented datasets}, i..e., $X_{l_i, a_j}$, $X_{a_j}$, $X_{l_i}$, $X_{all}$, and evaluate on all the test set covering both the in-group and the out-of-group test data, i.e, $X^{test}_{l_i, a_j}$, $X^{test}_{\neg l_i}$, $X^{test}_{\neg a_j}$,  and $X^{test}_{\neg l_i, \neg a_j}$. 

To further strengthen our analysis, we conduct the transferability experiment on the YueMotion dataset. We use the Cantonese elderly and Cantonese adults data on the YueMotion as the test set, and explore \ulcolor[darkgreen]{cross-group data augmentation} and \ulcolor[orange]{cross-group inference} for those test datasets. To measure the effectiveness of the cross-group transferability on YueMotion, we utilize the training datasets from BiMotion and compare the cross-group transfer methods with the baseline trained only on the YueMotion training dataset.

\textbf{Hyperparameters} We use the same hyperparameters in all experiments. For fine-tuning the model we use the following hyperparameter setting, i.e., AdamW optimizer~\cite{loshchilov2018decoupled} with a learning rate of 5e-5, a dropout rate of 0.1, a number of epochs of 20, and early stopping of 5 epochs.

\subsection{Evaluation Metrics}

Some speech utterances may consist of multiple emotion labels, e.g., happiness and surprise, which makes emotion recognition a multilabel problem. By framing the problem as multilabel, the occurrence of each emotion becomes sparse, thus leading to an imbalanced label distribution. To consider the imbalance distribution, we use binary F1-score to compute per-label evaluation performance and take the weighted F1-score over different emotion labels, which is an average of the F1 scores for each class weighted by the number of samples in that class. 



\section{Results and Discussion}

\begin{table}[!t]
    \caption{Evaluation results of \ulcolor[darkgreen]{cross-lingual and cross-age groups data augmentation} on BiMotion. \textbf{Cross-all} denotes cross-age cross-lingual data augmentation. \textbf{eng}, \textbf{zho}, \textbf{eld}, and \textbf{adu} denote English, Mandarin, elderly, and adults, respectively.}
    \label{tab:cga-result}
    \centering
    \resizebox{\linewidth}{!}{
        \begin{tabular}{c|c|c|c|c|c}
        \toprule
        \multirow{3}{*}[-0.3em]{\bf{\begin{tabular}[c]{@{}c@{}}Training\\Data\end{tabular}}} & \multicolumn{4}{c|}{\bf{Test Data}} & \multirow{3}{*}{\bf{Avg.}} \\ \cmidrule{2-5}
         & $\bm{l_i}$\bf{=eng} & $\bm{l_i}$\bf{=eng} & $\bm{l_i}$\bf{=zho} & $\bm{l_i}$\bf{=zho} & \\
         & $\bm{a_j}$\textbf{=eld} & $\bm{a_j}$\textbf{=adu} & $\bm{a_j}$\textbf{=eld} & $\bm{a_j}$\textbf{=adu} & \\ \midrule
        \textbf{Cross-all ($\bm{X^{trn}_{L, A}}$)} & \bf{68.06} & \bf{77.24} & 52.21 & \bf{97.40} & \textbf{73.73} \\
        \textbf{Cross-age ($\bm{X^{trn}_{l_i, A}}$)} & 66.96 & 75.47 & \bf{59.93} & 97.00 & 74.84 \\
        \textbf{Cross-lingual ($\bm{X^{trn}_{L, a_j}}$)} & 54.55 & 74.13 & 54.84 & 95.82 & 57.34 \\
        \textbf{Baseline ($\bm{X^{trn}_{l_i, a_j}}$)} & 55.12 & 70.28 & 57.00 & 97.00 & 69.85 \\
        \bottomrule
        \end{tabular}
    }
\end{table}

\subsection{Transferability via Cross-Group Inference}
\label{sec:cross-group-inference}

The results of our \ulcolor[orange]{cross-group inference} experiment are shown in Table~\ref{tab:cgi-result}. Based on our experiments, \ulcolor[orange]{cross-lingual, cross-age, and cross-age cross-lingual inferences} are not beneficial for improving the performance of the target group compared to the baseline trained on the specific language and age group, instead, they hinder the performance by a huge margin. For instance, the best \ulcolor[orange]{cross-age inference} comes from \textbf{English-Adults} to \textbf{English-Elderly}, however, it still hampers the performance of the in-group \textbf{English-Elderly} baseline by $\sim$15\% weighted F1-score. While for \ulcolor[orange]{cross-lingual inference}, the results are much worse with more than $\sim 30\%$ lower weighted F1-score than the corresponding in-group training baseline. This result suggests that there are some differences in speech features used to recognize emotion between different languages and different age groups, which makes transfer through \ulcolor[orange]{cross-group inference} not suitable for speech emotion recognition.

\begin{figure*}[!t]
    \centering
    \begin{minipage}{.28\linewidth}
        \centering
        \begingroup
        \includegraphics[width=\linewidth]{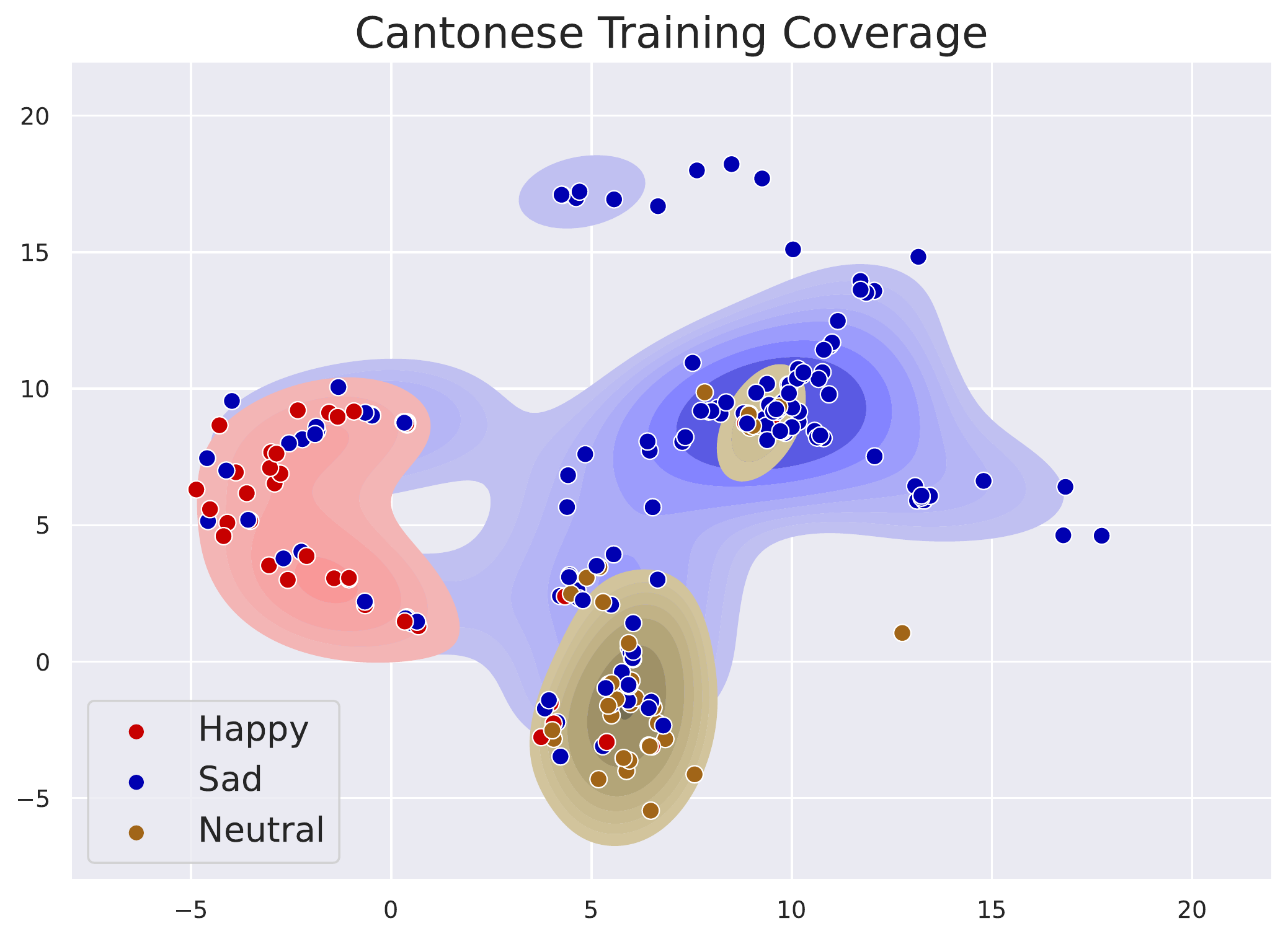}
        \endgroup
    \end{minipage}%
    \hspace{15pt}
    \begin{minipage}{.28\linewidth}
        \centering
        \begingroup
        \includegraphics[width=\linewidth]{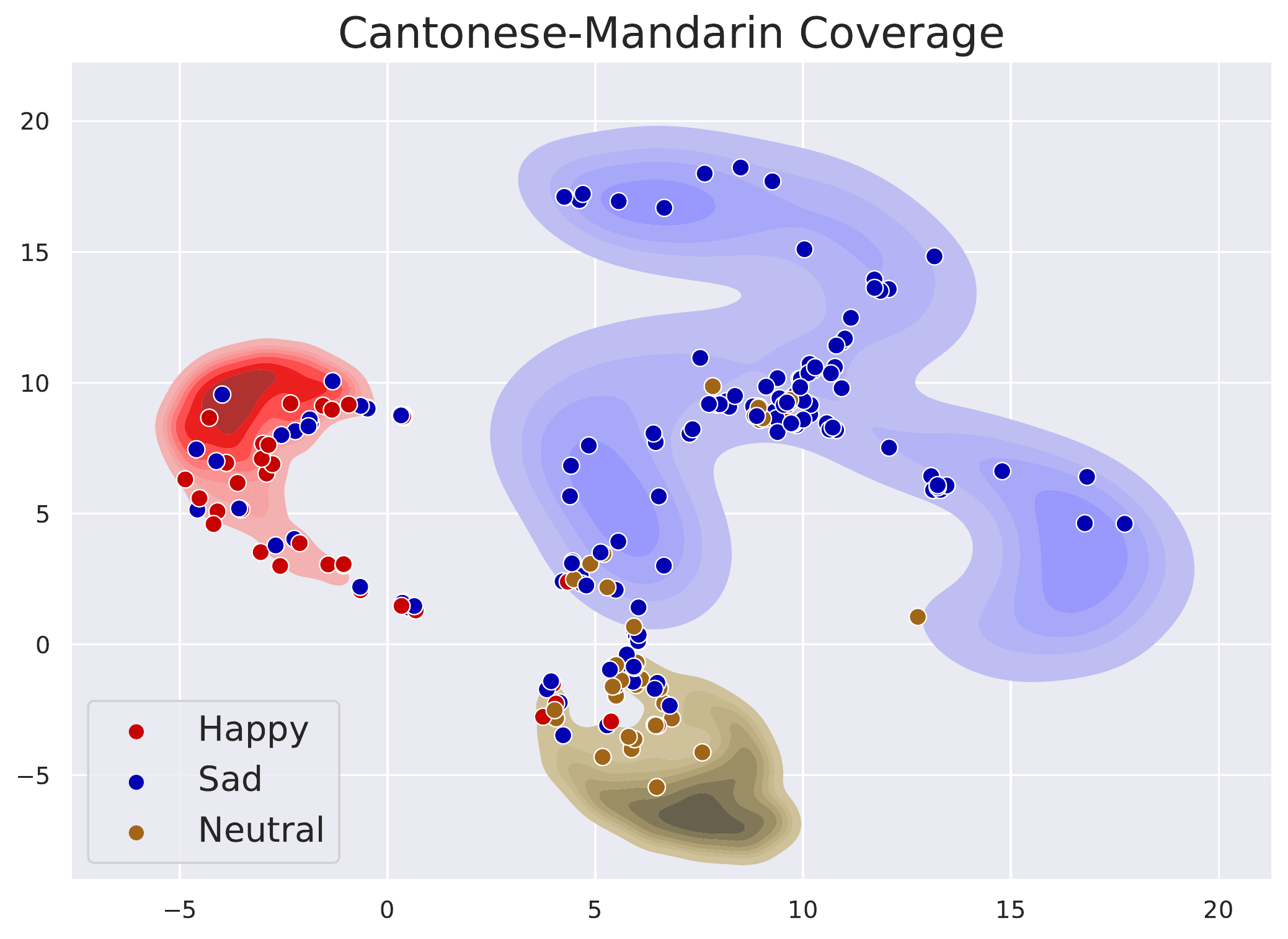}
        \endgroup
    \end{minipage}%
    \hspace{15pt}
    \begin{minipage}{.28\linewidth}
        \centering
        \begingroup
        \includegraphics[width=\linewidth]{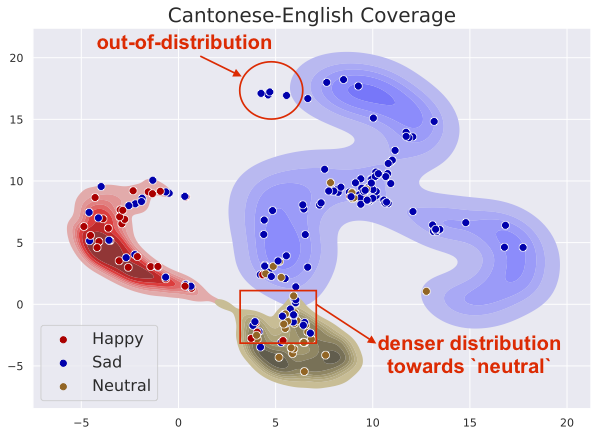}
        \endgroup
    \end{minipage}%
    \caption{\ulcolor[darkgreen]{Cross-group data augmentation} on the Cantonese-Adults data. \textcolor{blue}{Blue}, \textcolor{red}{red}, \textcolor{olive}{olive} regions represent the density plot of the training data for \textcolor{blue}{happy}, \textcolor{red}{sad}, and \textcolor{olive}{neutral} emotions, respectively.}
    \label{fig:aug-xlingual}
\end{figure*}

\begin{figure}[!t]
    \centering
    \begin{minipage}{.5\linewidth}
        \centering
        \begingroup
        \includegraphics[width=\linewidth]{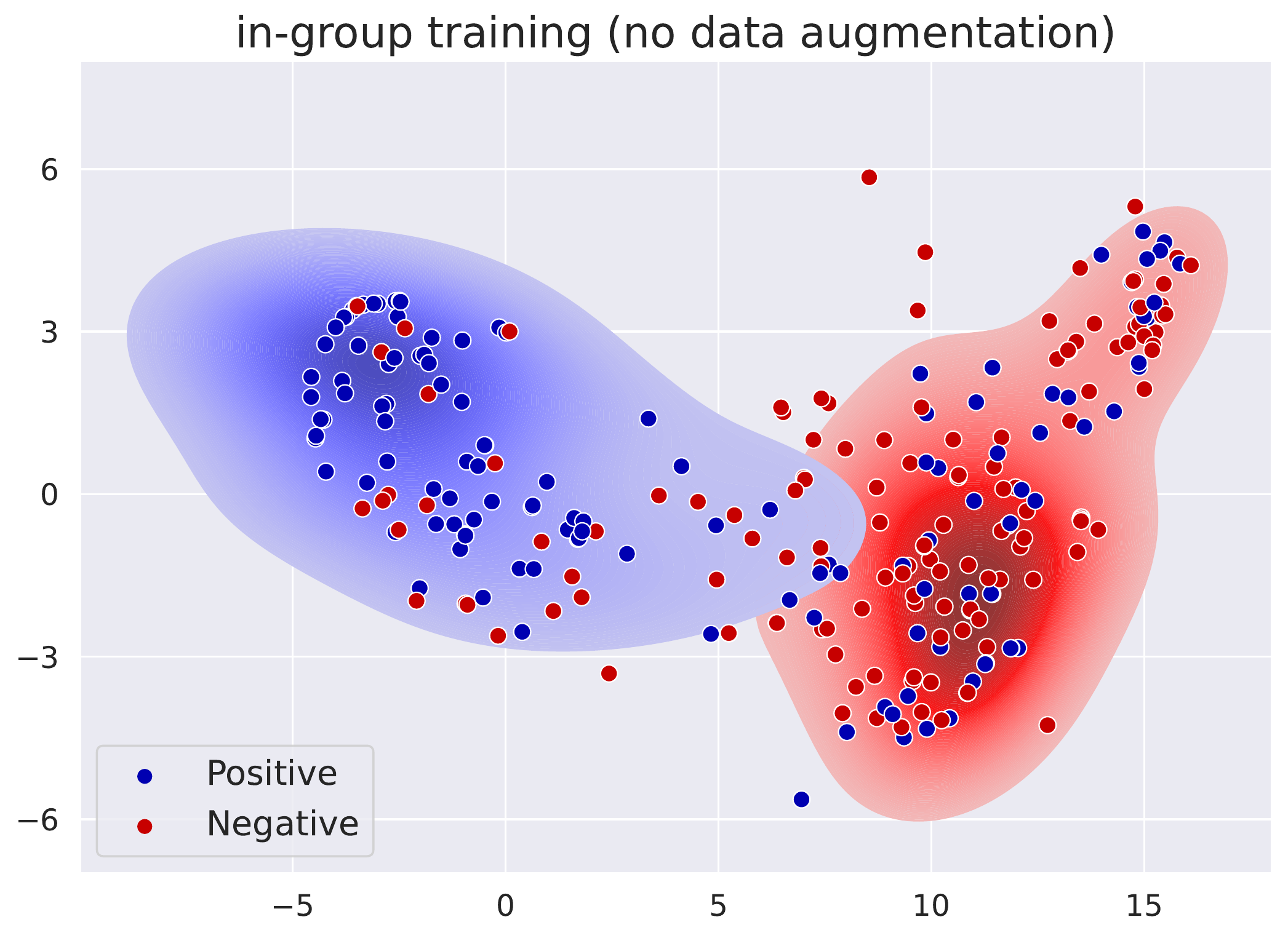}
        \endgroup
    \end{minipage}%
    \begin{minipage}{.5\linewidth}
        \centering
        \begingroup
        \includegraphics[width=\linewidth]{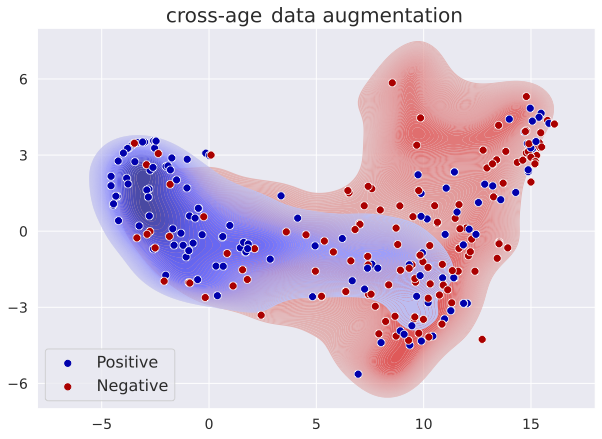}
        \endgroup
    \end{minipage}%
    \caption{\ulcolor[darkgreen]{Cross-group data augmentation} on the Mandarin-Elderly data. \textcolor{blue}{Blue}, and \textcolor{red}{red} regions denote the density of the training data for \textcolor{blue}{positive} and \textcolor{red}{negative} emotions, respectively.}
    \label{fig:aug-zh-eld}
    \vspace{-8pt}
\end{figure}

\subsection{Transferability via Cross-Group Data Augmentation}
\label{sec:cross-group-augmentation}

Despite having different characteristics, \ulcolor[darkgreen]{cross-group transfer through data augmentation} could provide richer and more diverse samples of emotion-induced speech and help to avoid overfitting problems, especially in a low-resource setting. Based on our cross-group data augmentation results in Table~\ref{tab:cga-result}, the \textbf{English-Elderly} and \textbf{English-Adults} groups benefit the most from the \ulcolor[darkgreen]{cross-group data augmentation}, gaining $\sim$12\% and $\sim$5\% weighted F1-score respectively. While for the Mandarin groups, the improvement is quite small. \textbf{Mandarin-Elderly} gains $\sim$3\% weighted F1-score with data augmentation from \textbf{Mandarin-Elderly}, while the result for the \textbf{Mandarin-Adults} remains the same, which is probably due to the limited data available from the \textbf{Mandarin-Elderly} group.

For \ulcolor[darkgreen]{cross-lingual data augmentation}, the effect is apparent with the only improvement coming from the \textbf{English-Adults} group, while the performance in other groups decreases. We conjecture this is due to the huge linguistic differences between English and Mandarin~\cite{lim2016cultural,fu2020speechsim}. Further analysis regarding the effect of language distance is discussed on \S\ref{sec:language-distance}.
Furthermore, a combination of \ulcolor[darkgreen]{cross-lingual and cross-age data augmentation} tends to improve the performance even higher. Specifically, the performance on \textbf{English-Elderly} and \textbf{English-Adults} increase by $\sim$13\% and $\sim$17\% weighted F1-score, respectively. The performance on \textbf{Mandarin-Adults} group improves marginally by $\sim$0.4\% weighted F1-score. On the other hand, the performance of \textbf{Mandarin-Elderly} decreases by $\sim$5\% weighted F1-score compared to the in-group training baseline. This might be caused by distributional shift due to the large amount of augmentation from other groups with respect to the amount of data in the \textbf{Mandarin-Elderly} group.

We further analyze the cross-group behavior further through feature-space projection using the model trained on all BiMotion training data. As shown in Figure~\ref{fig:aug-zh-eld}, \ulcolor[darkgreen]{cross-age data augmentation} improves the training coverage of the model resulting in a better generalization on unseen evaluation data, despite the speech feature differences among different age groups. This shows the potential of leveraging \ulcolor[darkgreen]{cross-age data augmentation} for modeling low-resource groups such as the elderly.

\subsection{Effect of Language Distance on Transferability}
\label{sec:language-distance}

We further analyze the effect of language distance on \ulcolor[darkgreen]{cross-lingual data augmentation}. Specifically, we explore the effect of \ulcolor[darkgreen]{cross-lingual data augmentation} from Mandarin and English to Cantonese. To balance the amount of data between English and Mandarin, we only utilize the ESD dataset~\cite{zhou2022esd} which contains 15K training, 1K validation, and 1.5K utterances for each English and Mandarin. For the Cantonese dataset, we utilize the newly constructed speech emotion dataset, YueMotion. As shown in Table~\ref{tab:yue-result}, the performance on the \textbf{Cantonese} data increases when the model is augmented with \textbf{Mandarin} data and decreases when the model is augmented with \textbf{English}. This follows the linguistic similarity of languages; Mandarin and Cantonese are more similar as both come from the same language family, i.e., Sino-Tibetan, compared to English that comes from the Indo-European language family. Similar pattern is also observed in the Mandarin test data when the Cantonese-Mandarin trained model outperforms the Cantonese-Mandarin-English model by $\sim$3\% weighted F1-score. 

We also analyze the cross-lingual behavior further through feature-space projection with the Cantonese-Mandarin-English trained model. As shown in Figure~\ref{fig:aug-xlingual}, the original Cantonese training data cannot cover all the test data. When Mandarin data is added, the training coverage improved which covers more test samples for each label. When English data is added, the training coverage is improved, but not without shifting the training distribution, e.g., the out-of-distribution on the sad label, causing the model to perform worse on the Cantonese evaluation data.

\begin{table}[!t]
    \caption{The effect of cross-lingual transferability in YueMotion.}
    \label{tab:yue-result}
    \centering
    \resizebox{0.85\linewidth}{!}{
    \begin{tabular}{c|c|c|c}
        \toprule
        \multirow{2}{*}[-0.2em]{\bf{\begin{tabular}[c]{@{}c@{}}Training\\Data\end{tabular}}} & \multicolumn{3}{c}{\bf{Test Data}} \\ \cmidrule{2-4}
         & \bf{$\bm{l_i}$=yue} & \bf{$\bm{l_i}$=eng} & \bf{$\bm{l_i}$=zho} \\
        \midrule
        \bf{Cross-lingual (yue+eng+zho)} & 47.59 & \textbf{92.61} & 93.75 \\
        \bf{Cross-lingual (yue+eng)} & 43.59 & 92.51 & - \\
        \bf{Cross-lingual (yue+zho)} & \textbf{51.60} & - & \textbf{97.00} \\
        \bf{Baseline (yue)} & 45.33 & - & - \\
        \bottomrule
    \end{tabular}
    }
\end{table}

\section{Conclusion}

We investigate the transferability of speech emotion recognition ability to achieve a better generalization for the elderly in low-resource languages. We construct an English-Mandarin speech emotion benchmark for adults and the elderly, namely BiMotion, from six publicly available datasets. We also construct, YueMotion, a low-resource speech emotion dataset for adults in Cantonese to analyze the effect of language distance in cross-lingual data augmentation. Based on the experiments, we conclude: 1) significantly distinct speech features are necessary to recognize emotion across different language and age groups, 2) although the speech features may vary across different groups, cross-group data augmentation is still beneficial to better generalize the model, and 3) language distance substantially affects the effectiveness of cross-lingual transferability.
Our analysis lays the groundwork for developing more effective speech emotion recognition models for low-resource groups, e.g., the elderly in low-resource languages.



\bibliographystyle{IEEEtran}
\bibliography{mybib}

\end{document}